\pdfoutput=1

\pdfoutput=1
\documentclass{article}
\usepackage{arxiv}

\newcommand{\name}{\texttt{PersonaTwin}\xspace}

\usepackage{easy-todo}
\usepackage{enumitem}
\usepackage{colortbl}
\usepackage{xcolor}
\usepackage{graphicx}
\usepackage{arydshln}
\usepackage{chngcntr}
\usepackage{mdframed}
\usepackage{tcolorbox}
\usepackage{multicol}
\usepackage{tikz}
\usetikzlibrary{automata, positioning, arrows.meta, shapes.geometric}
\usepackage{booktabs}
\usepackage{multirow}
\usepackage{anyfontsize}
\usepackage{pifont}   
\usepackage{threeparttable}
\usepackage{makecell}
\definecolor{Green}{RGB}{0,128,0}
\usepackage{hyperref}

\usepackage{tcolorbox}
\usepackage{listings}
\usepackage{xcolor}

\definecolor{codegreen}{rgb}{0,0.6,0}
\definecolor{codegray}{rgb}{0.5,0.5,0.5}
\definecolor{codepurple}{rgb}{0.58,0,0.82}
\definecolor{backcolour}{rgb}{0.95,0.95,0.92}

\lstdefinestyle{verilogStyle}{
    backgroundcolor=\color{backcolour},   
    commentstyle=\color{codegreen},
    keywordstyle=\color{blue},
    numberstyle=\tiny\color{codegray},
    stringstyle=\color{codepurple},
    basicstyle=\ttfamily\footnotesize,
    breakatwhitespace=false,         
    breaklines=true,                 
    captionpos=b,                    
    keepspaces=true,                 
    numbers=left,                    
    numbersep=5pt,                  
    showspaces=false,                
    showstringspaces=false,
    showtabs=false,                  
    tabsize=2
}

\tcbuselibrary{listingsutf8}
\newtcblisting{verilogcode}{
    colback=white,
    colframe=black,
    listing only,
    listing options={style=verilogStyle, language=Verilog},
    left=5pt, right=5pt, top=5pt, bottom=5pt
}

\usepackage{amsmath, amssymb, mathtools, amsthm}
\usepackage{titlesec}
\usepackage[capitalize,noabbrev]{cleveref}
\usepackage{booktabs}

\usepackage{times}
\usepackage{latexsym}
\usepackage[T1]{fontenc}
\usepackage[utf8]{inputenc}
\usepackage{microtype}
\usepackage{inconsolata}
\usepackage[linesnumbered,ruled,vlined]{algorithm2e}
\usepackage{amsmath}
\usepackage{natbib}

\renewcommand\cite{\citep}	

\title{PersonaTwin: A Multi-Tier Prompt Conditioning Framework for Generating and Evaluating Personalized Digital Twins}

\author{Sihan Chen,$^1$ John P. Lalor,$^{2,3}$ Yi Yang,$^4$ Ahmed Abbasi$^{2,3}$ \\
$^1$Viterbi School of Engineering, University of Southern California\\
$^2$Human-centered Analytics Lab, University of Notre Dame\\
$^3$Department of IT, Analytics, and Operations, University of Notre Dame\\
$^4$Department of Information Systems, Business Statistics and Operations Management, HKUST\\
{\small\texttt{schen976@usc.edu}, \texttt{john.lalor@nd.edu}, \texttt{imyiyang@ust.hk}, \texttt{aabbasi@nd.edu} }
}

\begin{document}
\maketitle
\begin{abstract}

While large language models (LLMs) afford new possibilities for user modeling and approximation of human behaviors, they often fail to capture the multidimensional nuances of individual users. In this work, we introduce \name, a multi-tier prompt conditioning framework that builds adaptive digital twins by integrating demographic, behavioral, and psychometric data. Using a comprehensive data set in the healthcare context of more than 8,500 individuals, we systematically benchmark \name against standard LLM outputs, and our rigorous evaluation unites state-of-the-art text similarity metrics with dedicated demographic parity assessments, ensuring that generated responses remain accurate and unbiased. Experimental results show that our framework produces simulation fidelity on par with oracle settings. Moreover, downstream models trained on persona-twins approximate models trained on individuals in terms of prediction and fairness metrics across both GPT-4o-based and Llama-based models. Together, these findings underscore  the potential for LLM digital twin-based approaches in producing realistic and emotionally nuanced user simulations, offering a powerful tool for personalized digital user modeling and behavior analysis.

\end{abstract}

\section{Introduction}

\label{sec:intro}
Large language models (LLMs) present exciting opportunities for user modeling, behavior analysis, and understanding and improving the human condition. Opportunities abound across an array of contexts including healthcare, education, etc. For instance, a pressing healthcare challenge is the development of conversational systems that truly account for the nuanced
experiences and identities of individual patients
\cite{davenport2019potential,jiang2017artificial}. In telemedicine or mental
health coaching scenarios, clinicians require tools that adapt dynamically to
each patient’s demographic, behavioral, and psychological profile, rather than
offering generic responses. Although large language models such as GPT-4
\cite{openai2023gpt4} and Llama-3-70b \cite{Grattafiori2024llama3} have shown
substantial improvement in natural language processing tasks--demonstrated by
benchmarks in medical QA datasets, automated note taking, and patient triage use
cases--they still struggle to model the multifaceted nature of personal identity
in real world settings \cite{articleLaranjo}. Numerous persona-based
conversational frameworks have begun to address this gap by incorporating basic
user attributes into language models, and studies have demonstrated modest gains in engagement and trust when even minimal demographic cues are included \cite{inproceedingsqual}. However, many of these frameworks remain limited by static or
simplistic representations that fail to capture evolving factors. 
In the health setting, for instance, these frameworks fail to represent health behaviors over time, emotional states during stressful events, and shifting attitudes toward medical professionals \cite{huang2024leveragingfinitestatesemotion, Guo2024HOAC}.

To overcome these limitations, we draw on the concept of digital twins,
originally popularized in engineering, to represent physical systems virtually 
\cite{grieves2014digital}.
In our adaptation, a ``digital twin'' for conversational AI is a virtual replica
of a user (e.g., a patient) that encapsulates not only demographic information (e.g., age,
gender, and socioeconomic status), but also behavioral data (e.g., physical
activity, dialogue habits, and compliance with medications), along with
psychological attributes (e.g., anxiety levels, trust, and perceived literacy) \cite{meijer2023digital,lukaniszyn2024digital}. 
This framework is particularly relevant in scenarios such as mental health chatbots or chronic disease management systems, where the emotional and psychological realism of the dialogue can directly impact patient adherence and satisfaction. 

However, creating such multidimensional and adaptive representations raises
several methodological hurdles. First, many existing language-based approaches
provide only a narrow view of a user’s identity, focusing predominantly
on stylistic or linguistic features while neglecting deeper demographic or
psychometric attributes. Second, static systems do not adapt to shifting
contexts, including new symptoms or a gradual erosion of trust, resulting in repetitive or misaligned conversations. Third, there is
a lack of comprehensive evaluation benchmarks that jointly measure
factual correctness, emotional coherence, and alignment with actual user
expressions. For instance, in clinical contexts, much of the previous NLP work has focused on factual
accuracy, leaving emotional nuance and user alignment underexplored
\cite{jiang2017artificial}.

To address these challenges, we introduce \name, a multi-tier prompt conditioning framework that systematically integrates demographic, behavioral, and psychological data into a comprehensive digital twin. Our approach employs a structured methodology in which each level of user information is processed and encoded into the model prompt \cite{DBLP:journals/corr/abs-2104-08691,chen2024recenttrendspersonalizeddialogue}. \name consists of two parts, \textbf{Multi-tiered Conditioning for Digital Twin Creation} and \textbf{Conversation Update Loop}. In the first part, step 1 involves mapping person-level persona metadata to persona information tiers such as demographics, behavioral, and psychological information; whereas step 2 initializes the digital twin. In the second part, the instantiated digital twin is iteratively updated with the real person's previous conversational responses to psychometric questions (e.g., related to numeracy, anxiety, etc.). This layered technique enables the model to produce simulated dialogues that are not only contextually relevant but also capable of reflecting shifting user states as new data are introduced \cite{reimers2019sentence}. We tested our framework using a large-scale psychometric dataset of more than 8,500
respondents~\cite{abbasi-etal-2021-constructing}, which provides a rich combination of survey-based measures, user-generated text, and demographic information. By incorporating real responses on health numeracy, medical visit anxiety, and trust in healthcare providers, we ensure that our simulations reflect authentic user experiences while maintaining privacy through deidentification and ethical safeguards \cite{cascella2023evaluating}.

To rigorously evaluate \name, we implemented a dual-pronged strategy. First, we employ state-of-the-art text similarity metrics
to measure how closely the digital twin-generated output
matches the actual user responses 
\cite{reimers2019sentence,song2020mpnetmaskedpermutedpretraining,wang2020minilm}. 
Second, we use a downstream NLP prediction task to examine the efficacy of the generated twins, relative to the actual users, in terms of the fine-tuned model's predictive performance and fairness assessments across key demographic dimensions \cite{DBLP:journals/corr/HardtPS16,barocas2019fairness}.

Our key contributions are: (i) we introduce \name, a multi-tier framework that integrates demographic, behavioral, and psychological data to generate adaptive digital twins, enhancing realism with LLM-driven personal insights, (ii) we generate 8,500+ digital twins representing diverse personas and validate response fidelity using conditioned experiments and advanced similarity metrics, and (iii) we conduct a rigorous downstream evaluation of models trained/tested on generated personas versus actual users and show that the persona-based models achieve comparable predictive power and fairness outcomes.\footnote{Our code is available on GitHub: \url{https://github.com/nd-hal/psych-agent-llm}.}

\section{Related Work}
\subsection{Simulative Persona Construction and the Importance of Digital Twins}

A pioneering study by \citet{park2023generativeagentsinteractivesimulacra} laid the groundwork for persona-based conversational systems by simulating a small town of 25 virtual characters using simplified models of human cognition to enable dialogue. 
Recent advances in generative agents have begun to explore the ability of LLMs to emulate more precise human behaviors. For instance, \citet{park2024generativeagentsimulations1000} simulate survey responses for 1,000 individuals based on audio interviews with participants. Similarly, \citet{xu2024theagentcompanybenchmarkingllmagents} benchmark LLM agents on consequential real-world tasks. In parallel, \citet{chuang-etal-2024-beyond} develop digital twins using a belief network to capture open-domain dimensions--such as those revealed in the Controversial Beliefs Survey--broadening the scope of persona construction. Moreover, \citet{shao-etal-2023-character} propose Character-LLM, an approach that crawls online records and stories of historical or fictional figures to serve as persona inputs, thereby enriching the contextual and experiential background of the simulated agents. Additionally, as discussed in \citep{meister2024benchmarking}, “steering methods” offer promising strategies to guide the behavior of simulated agents. However, these studies often face two major challenges: (1) some approaches rely solely on unstructured text inputs yet lack the precise control needed to ensure consistency in the perspectives from which user content is drawn; and (2) other methods incorporate structured data, but primarily focus on personal background without delving deeply into the internal psychological traits and behavioral dynamics of individuals.

Furthermore, \citet{salemi2024lamp} introduce LaMP, a comprehensive benchmark and retrieval-augmentation framework that conditions LLMs on fine-grained user profiles--spanning classification and generation tasks--to produce personalized outputs, demonstrating significant gains in both zero-shot and fine-tuned settings. Meanwhile, \citet{sorokovikova2024llms} provide empirical evidence that LLMs (e.g.\ Llama-2, GPT-4, Mixtral) can simulate stable Big Five personality traits, revealing the potential of LLM-driven agents to model intra-individual psychological characteristics with consistency across varied prompts.

Our work addresses this challenge by taking a fine-grained, high-dimensional approach to simulating individual personas. We integrate psychological, behavioral, personal background, and linguistic style information to construct digital twins that capture the nuanced and evolving nature of real human identities. By leveraging authentic user inputs as benchmarks, our framework explores replication of core behavioral patterns and  individual variability that is typically lost in more simplistic, one-dimensional models. We demonstrate the potential for enriched representations for generating digital twins that better reflect real human behavior.

\subsection{Evaluation Metrics for Fairness and Authenticity in Generative Agents}

Evaluating generative agents requires robust metrics that capture not only the linguistic quality but also the downstream efficacy and fairness of the generated responses. Many studies have adopted LLM-based evaluation methods, either by leveraging off-the-shelf or fine-tuned LLMs, or by incorporating human evaluators, to assess the authenticity of generated text \cite{jandaghi-etal-2024-faithful, mendonca-etal-2024-soda, park2023generativeagentsinteractivesimulacra, chiang-lee-2023-large}. Although these methods have demonstrated promising results, they are not without drawbacks. In scenarios involving large volumes of language data, extensive human evaluation quickly becomes both cost- and time-inefficient. Furthermore, the performance of these evaluation frameworks relies heavily on the underlying LLMs, which may harbor inherent biases or produce unpredictable outputs \cite{lin-chen-2023-llm}. Moreover, traditional automatic metrics such as BLEU, ROUGE, METEOR, and CIDEr \cite{papineni-etal-2002-bleu, lin-2004-rouge, banerjee-lavie-2005-meteor, oliveira-dos-santos-etal-2021-cider} often fall short in capturing deeper semantic alignment and social fairness \cite{zhang2020bertscoreevaluatingtextgeneration}.

To overcome these challenges, embedding-based metrics, particularly those leveraging BERT, have emerged as a promising balance between effectiveness and efficiency. For example, BERTScore \cite{zhang2020bertscoreevaluatingtextgeneration} computes semantic similarity by comparing contextual embeddings of generated texts with those of reference texts, thereby capturing nuances that traditional n-gram metrics often miss. Moreover, Zhu and Bhat \cite{zhu-bhat-2020-gruen} introduce GRUEN, a reference-less framework that leverages a BERT-based model to reliably assess the linguistic quality of generated text. Additionally, several studies have extended BERT-based evaluations beyond mere semantic alignment. For instance, \citet{lalor-etal-2022-benchmarking} fine-tuned BERT and RoBERTa models and assessed fairness via disparate impact scores across multiple demographic attributes. These applications underscore how BERT and its variants can provide a robust and efficient framework for evaluating both the authenticity and fairness of generative agents, offering a viable alternative to more resource-intensive LLM-based or human evaluation strategies \cite{lin-chen-2023-llm}.

\section{Methodology}

\label{sec:methodology}

In this section, we introduce the structure of our proposed framework, \name (\S \ref{subsec:construction}), and then detailed our evaluation metrics (\S \ref{ssec:eval}). 

\begin{figure*}[h!]
    \centering
    \includegraphics[width=1\linewidth]{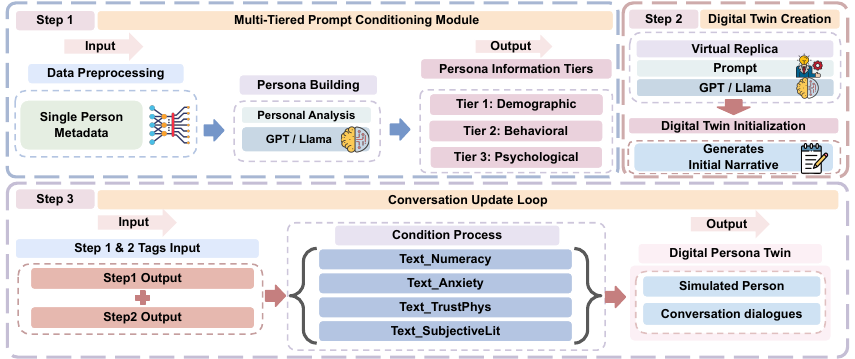}
    \caption{An overview of the \name framework, including 1) Multi-Tiered Prompt Conditioning Module, 2) Digital Twin Creation, and 3) Conversation Update Loop.}
    \label{fig:overview}
    \vspace{-7pt}
\end{figure*}

\subsection{Digital Twin Construction}
\label{subsec:construction}

In this section, we detail our two-stage methodology for constructing
and refining digital twins using large language models (LLMs). We denote our framework by \name. In Stage~1, we create an initial digital twin by
integrating multidimensional user data into a structured prompt for LLM. In Stage~2, we iteratively update the digital twin based on new user input and conversation data, thus capturing temporal changes in user states. The overall process for constructing and refining digital twins is formally detailed in Figure~\ref{fig:overview}.

\subsubsection{Digital Twin Initialization}
\label{subsubsec:instance_init}

For the first step of initializing digital twins, we systematically collect and preprocess
heterogeneous user data: $D = \{d_1, d_2, \dots, d_N\}$,
where each \(d_i\) can represent a demographic (age, race, income), behavioral
(physical activity, dietary habits, medication adherence), or psychological
(trust, anxiety levels, literacy, numeracy) attributes. A
preprocessing function \( I(\cdot) \) converts and normalizes all these inputs
into a structured representation:

\begin{equation}
X = I(D) = \bigl[X_{\text{dem}},\, X_{\text{beh}},\, X_{\text{psy}}\bigr].
\end{equation}

Here, \(X_{\text{dem}}, X_{\text{beh}}, X_{\text{psy}}\), respectively, encode
the demographic, behavioral, and psychological data in vectorized or categorical
form.\footnote{Refer to Appendix \ref{sec:furtherdetails} for further details.}

\paragraph{Multi-Tiered Template Functions.}
Unlike a simple concatenation of all features, \name\ employs three dedicated
template functions: \texttt{Template\_dem}, \texttt{Template\_beh}, and
\texttt{Template\_psy}, each tuned to capture domain-specific nuances. These
functions provide additional context such as causal phrases (e.g., ``Because the person has high anxiety...''), relevant guidelines, or rhetorical
questions that nudge the LLM to infuse the output with emotional tone and
factual correctness.

Formally,
\begin{equation}
\begin{aligned}
P_{\text{dem}} &= \texttt{Template\_dem}(X_{\text{dem}}),\\
P_{\text{beh}} &= \texttt{Template\_beh}(X_{\text{beh}}),\\
P_{\text{psy}} &= \texttt{Template\_psy}(X_{\text{psy}}).
\end{aligned}
\end{equation}

where each template can \emph{rewrite, summarize, or highlight} the most
critical aspects of the data. For example, if \(X_{\text{psy}}\) indicates a
high anxiety level, \texttt{Template\_psy} might produce text emphasizing the
user's tendency to worry about medical procedures, thus improving emotional
realism. 

\paragraph{Initial Digital Twin Generation.}
We concatenate the tier-specific prompts to form a composite prompt:
\begin{equation}
P = \text{Concat}(P_{\text{dem}}, P_{\text{beh}}, P_{\text{psy}}),
\end{equation}
which is passed to a selected LLM \(G(\cdot)\) to obtain the initial digital
twin \(T_{0}\):
\begin{equation}
T_{0} = G(P).
\end{equation}
This \emph{initialization} step produces a coherent user narrative or persona
that encapsulates the baseline demographic, behavioral, and psychological
characteristics. 

\subsubsection{Conversation Data Integration and Dynamic Update Loop}
\label{subsubsec:conv_integration}

Although the initial digital twin \(T_{0}\) provides a rich snapshot of the
focal user, it cannot reflect changes in user states or additional data acquired
over time. This motivates our second stage, where we iteratively integrate user's conversations (e.g., with psychiatrists) into our digital twin framework.

\paragraph{Conversation Update Mechanism.}
At each iteration \(t\), the user query \(Q_{t}\) corresponds to one of the four
types of prompts in Table~\ref{tab:questions} (i.e., \texttt{Text\_Numeracy},
\texttt{Text\_Anxiety}, \texttt{Text\_TrustPhys}, or
\texttt{Text\_SubjectiveLit}). We obtain a corresponding user response
\(R_{t}\), which may be drawn from real user data or a newly simulated input. An
update function \(U\) refines the digital twin \(T_{t}\) as follows:
\begin{equation}
    T_{t+1} = U(T_{t}, Q_{t}, R_{t}).
\end{equation}

In practice, \(U\) rechecks each prompt template to integrate relevant changes.
For example, if \(R_{t}\) indicates a dose increase for a medication,
\texttt{Template\_beh} is updated to reflect this new regimen. In contrast, if
the user contradicts an earlier statement (e.g., previously denied smoking but
now mentions occasional use), \texttt{Template\_beh} reconciles these by
prioritizing the recent self-report while tagging older statements as ``possible
past data.'' This conflict resolution policy ensures that the most up-to-date
information prevails, although older data are retained for longitudinal context.

\paragraph{Multi-Tiered Prompt Conditioning Experiments.}
Rather than simply updating static persona templates, we devised eight
distinct sub-sample conditions, denoted by 
\begin{equation}
{T}' = \{T'_1, T'_2, \dots, T'_8\}.
\end{equation}
to assess how well \name generates realistic user responses under varying
degrees of known personal and conversational information pertaining to the focal user. These conditions are based on two factors: (1) whether the simulated person receives their paired users' three persona information tiers (i.e., demographic, behavioral, psychological); (2) whether the simulated person receives some or none of the four potential conversation updates (called few-shot if yes, zero-shot if no).   

Specifically, we define \(T'_1\)
as \textbf{\emph{Persona Oracle}}, where the system prompt includes persona information tier data and the all four conversation updates are revealed,
thus serving as a maximum informed oracle. We then introduce four
\textbf{\emph{Persona Few-shot}} variants, \(T'_2, T'_3, T'_4,\) and \(T'_5\), each
withholding one of the four real responses to test the model’s ability to infer
missing content from partial context. Next, \(T'_6\), labeled
\textbf{\emph{Persona Zero-shot}}, omits all real answers entirely, requiring
the LLM to generate plausible responses purely from the user's personal
attributes. In contrast, \(T'_7\), named \textbf{\emph{Few-shot Oracle}},
removes all demographic and behavioral cues but supplies the actual four
responses, allowing the model to ground its simulation in user statements while
lacking direct persona data. Finally, \(T'_8\), the \textbf{\emph{Zero-shot}} condition, excludes both personal information and true answers,
evaluating how the model performs with virtually no contextual cues. Evaluating
each digital twin across \(\mathcal{T}'\) and the four queries in
Table~\ref{tab:questions} allows us to gauge the influence of different 
configurations on the coherence, precision and consistency of the simulated
person responses.

\begin{table}[ht]
\centering
\small
\setlength{\tabcolsep}{4pt}
\renewcommand{\arraystretch}{1.5}
\caption{Q\&A Prompts for Digital Twin Updates}
\label{tab:questions}
\begin{tabular}{@{}p{3cm}p{0.55\linewidth}@{}}
\toprule
\textbf{Question Dimension} & \textbf{Prompt} \\ \midrule
\textbf{Numeracy} & \textit{``In a few sentences, please describe an experience in your life that demonstrated your knowledge of health or medical issues.''} \\
\textbf{Anxiety}  & \textit{``In a few sentences, please describe what makes you feel most anxious or worried when visiting the doctor's office.''} \\
\textbf{Trust in Physician} & \textit{``In a few sentences, please explain the reasons why you trust or distrust your primary care physician. If you do not have a primary care physician, please answer in regard to doctors in general.''}
\\
\textbf{Subjective Health\newline Literacy} & \textit{``In a few sentences, please describe to what degree do you feel you have the capacity to obtain, process, and understand basic health information and services needed to make appropriate health decisions?''} \\ 
\bottomrule
\end{tabular}
\end{table}

\subsection{Evaluation Metrics}
\label{ssec:eval}

\subsubsection{Simulated Person Response Similarity}
Let \(t\) be a text document (e.g., a patient response), and let \(\displaystyle
f: \mathcal{T} \rightarrow \mathbb{R}^d\) be an embedding function provided by a
pre-trained language model such as \texttt{BERT\_CLS}, \texttt{MiniLM-L6-v2}, or
\texttt{mpnet-base-v2}. For any text \(t\), we obtain its embedding vector
\(\mathbf{v}\) via $\mathbf{v} = f(t)$.

In our setting, each user is asked one of four domain-specific questions pertaining to a specific health dimension (Table~\ref{tab:questions}). 
Let \(t_{\text{gen}}(q)\) be the LLM-generated
response and \(t_{\text{true}}(q)\) the corresponding ground-truth response for
question \(q\). We map each to its embedding space, yielding
\begin{align}    
    \mathbf{v}_{\text{gen}}(q) &= f\bigl(t_{\text{gen}}(q)\bigr)\\
    \mathbf{v}_{\text{true}}(q) &= f\bigl(t_{\text{true}}(q)\bigr)
\end{align}

We then compute their cosine similarity,
\begin{equation}
  \text{sim}\bigl(t_{\text{gen}}(q), t_{\text{true}}(q)\bigr)
  \;=\;
  \frac{\langle \mathbf{v}_{\text{gen}}(q),\,\mathbf{v}_{\text{true}}(q)\rangle}
  {\|\mathbf{v}_{\text{gen}}(q)\|\;\|\mathbf{v}_{\text{true}}(q)\|},
\end{equation}
where \(\langle \cdot,\cdot\rangle\) denotes the dot product, and \(\|\cdot\|\)
denotes the Euclidean norm. This similarity measure lies in the interval
\(\bigl[-1,1\bigr]\), with higher values indicating stronger alignment between
the generated response and the ground-truth text.

\subsubsection{Downstream Prediction and Fairness}

We assess the downstream prediction power and fairness of models fine-tuned using the simulated persona-twins versus actual users by drawing on the methodology described in \citet{lalor-etal-2022-benchmarking}, which focuses on quantifying prediction metrics and related intersectional biases across multiple demographic dimensions. Our evaluation framework includes the following components:

\textbf{Model Fine-Tuning and Hyperparameter Settings.}  
   We fine-tuned BERT model for five epochs using a batch size of 32, a learning rate of 1e-5, and a weight decay of 0.01. The model that achieved the lowest validation loss was saved as the final model. This approach balances training quality and overfitting prevention. For each experimental setting, we conducted five-fold cross validation.

 \textbf{Performance and Fairness Metrics.}  
   In addition to standard performance metrics such as AUC, F1 score, mean squared error (MSE), and Pearson's correlation coefficient, we evaluated fairness using a series of disparate impact (DI) metrics. Specifically, DI scores were computed for individual demographic attributes: age, gender, race, education, and income, as well as for their intersectional combinations. These metrics help to reveal any biases in model predictions across different subgroups. 

Collectively, the downstream prediction task is intended to highlight the inference potential and fairness of models trained on the constructed digital persona twins relative to the actual users.

\section{Experiments}

\begin{table*}[ht!]
\setlength{\tabcolsep}{3pt} 
\centering
\resizebox{1\textwidth}{!}{ \fontsize{6}{7}\selectfont
\begin{tabular}{lcccccccccccc}
\toprule
 &
\multicolumn{4}{c}{\textbf{bert\_CLS}} &
\multicolumn{4}{c}{\textbf{sbert\_MiniLM}} &
\multicolumn{4}{c}{\textbf{sbert\_mpnet}} \\
\textbf{Condition} & \textbf{Anxiety} & \textbf{Numeracy} & \textbf{Lit} & \textbf{TrustPhys} &
\textbf{Anxiety} & \textbf{Numeracy} & \textbf{Lit} & \textbf{TrustPhys} &
\textbf{Anxiety} & \textbf{Numeracy} & \textbf{Lit} & \textbf{TrustPhys} \\
\midrule
\multicolumn{13}{c}{\textbf{GPT-4o}}\\
\textbf{Persona Oracle} & \textbf{0.952} & 0.952 & \textbf{0.970} &\textbf{0.965} & \textbf{0.535} & \textbf{0.291}
& 0.586 & \textbf{0.589} & \textbf{0.599} & \textbf{0.361} & \textbf{0.647} & \textbf{0.683} \\
\textbf{Few-Shot Oracle} & 0.946 & 0.951 & 0.968 & 0.962 & 0.504 & 0.285 & \textbf{0.587} &
0.562 & 0.575 & 0.354 & 0.644 & 0.660 \\
\textbf{Persona Few-shot} & 0.949* & \textbf{0.953*} & 0.968* & 0.961* & 0.490 & 0.272* & 0.553 & 0.536*
& 0.556 & 0.337* & 0.620* & 0.641* \\
\textbf{Persona Zero-shot} & 0.939* & 0.943 & 0.964* & 0.952 & 0.491 & 0.227 & 0.500 &
0.515 & 0.554 & 0.292 & 0.582 & 0.624 \\
\textbf{Zero-Shot} & 0.937 & 0.942 & 0.962 & 0.954 & 0.492 & 0.240 & 0.553 &
0.513 & 0.562 & 0.299 & 0.612 & 0.620 \\
\midrule
\multicolumn{13}{c}{\textbf{Llama-3-70b}}\\
\textbf{Persona Oracle} & \textbf{0.957} & \textbf{0.959} & \textbf{0.971} & \textbf{0.961} & \textbf{0.526} & 0.325 & \textbf{0.571}
& \textbf{0.600} & \textbf{0.600} & 0.383 & \textbf{0.615} & \textbf{0.689} \\
\textbf{Few-Shot Oracle} & 0.955 & 0.958 & \textbf{0.971} & 0.960 & 0.510 & \textbf{0.330} & 0.564 &
0.593 & 0.582 & \textbf{0.385} & 0.604 & 0.683 \\
\textbf{Persona Few-shot} & 0.955* & 0.956* & 0.969* & 0.956* & 0.486* & 0.291* & 0.544* & 0.545*
& 0.555* & 0.346 & 0.595* & 0.650* \\
\textbf{Persona Zero-shot} & 0.941 & 0.949* & 0.966 & 0.956* & 0.476 & 0.282* & 0.517* &
0.506 & 0.533* & 0.327 & 0.577* & 0.623* \\
\textbf{Zero-Shot} & 0.931 & 0.942 & 0.967 & 0.950 & 0.476 & 0.277 & 0.510 &
0.503 & 0.522 & 0.306 & 0.533 & 0.609 \\
\bottomrule
\end{tabular}
}
\caption{Similarity scores for GPT-4o (top) and Llama-3-70b (bottom) models across different conditions.\newline $*$ indicates similarity scores significantly higher than the zero-shot baseline ($p<0.05$).}
\label{tab:similarity_scores_4o}
\end{table*}

\subsection{Datasets}
For this study, we utilized the psychometric dataset from 
\citet{abbasi-etal-2021-constructing}. 
The dataset
comprises survey-based psychometric measures alongside user-generated text,
gathered from over 8,500 respondents. The primary psychometric dimensions
measured include trust in physicians, anxiety visiting the doctor’s office,
health numeracy, and subjective health literacy. These dimensions are critical
to understanding user behavior in healthcare and were selected based on their
relevance to an array of health outcomes. The English-language dataset offers a rich blend of
structured survey responses and unstructured text, including detailed
demographic information (e.g., age, gender, race, education, and income) alongside
psychometric and behavior measures. This enables a comprehensive analysis of how human factors influence text-based responses
\citep[e.g.,][]{zhou2023causal,gohar2023survey,dai2024mitigate,van2024undesirable}.\footnote{The data collection protocol for \citep{abbas-lichouri-2021-tpt} was approved by the University of Virginia IRB-SBS under SBS Number 2017014300.}

\subsection{Models}

To generate the simulated responses \(t_{\text{gen}}\), we employ two LLMs:
\texttt{GPT-4o} and \texttt{Llama-3-70b}.\footnote{Refer to Appendix \ref{ssec:hyperparams} for implementation details.} 
We then use each of the three
pre-trained models (\texttt{bert-base-uncased}, \texttt{MiniLM-L6-v2}, and
\texttt{mpnet-base-v2}) as embedding functions \(f\) to assess the quality of
the generated text from multiple representational perspectives \citep{reimers2019sentence}. 
This way, we evaluate how faithfully the model output matches the user's
actual responses, and also examine the robustness of our similarity scores
to variations in the underlying embedding space.
    
\begin{figure*}[htbp]

    \centering
    \includegraphics[width=0.9\textwidth]{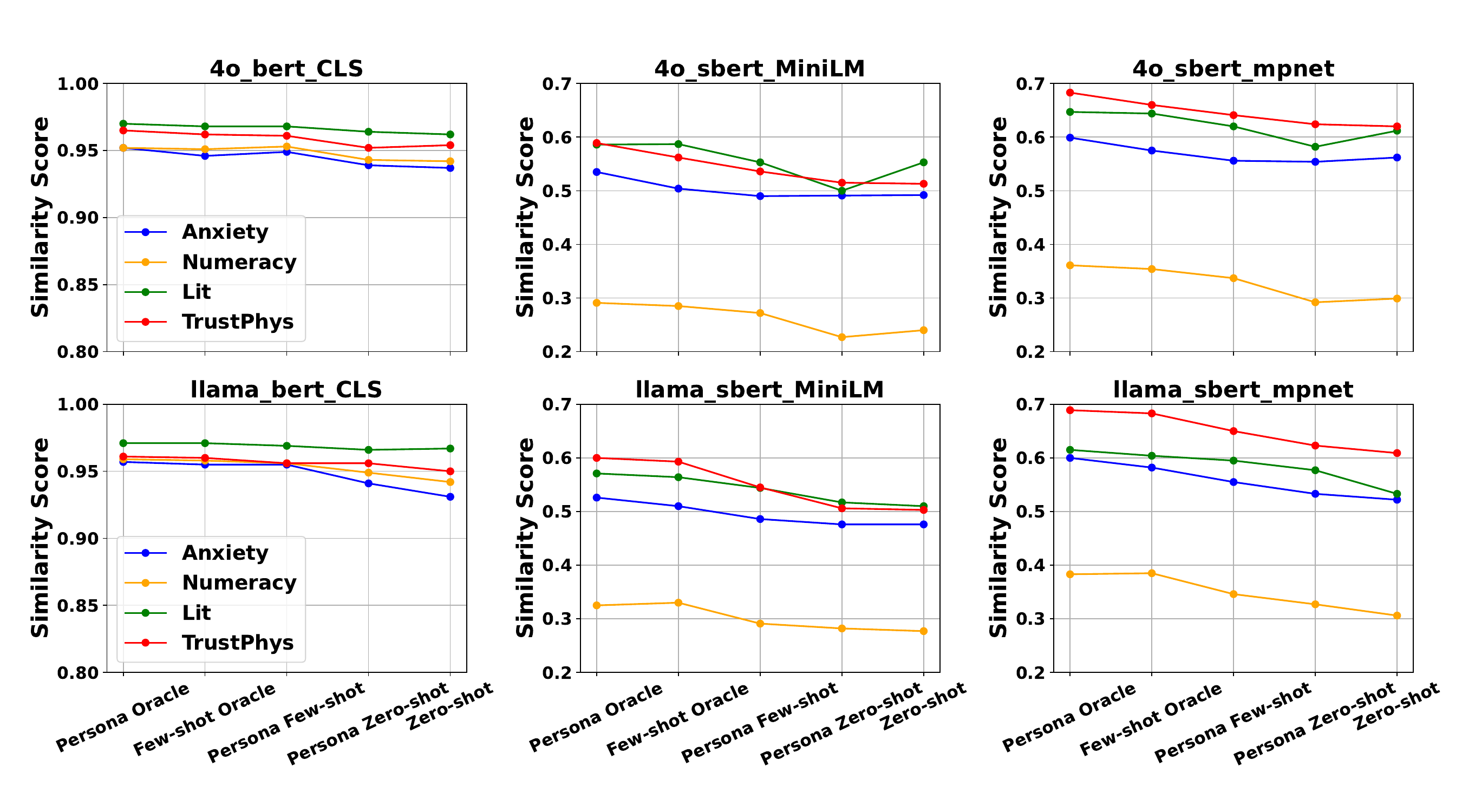}
    \vspace{-14pt}
    \caption{Comparison of similarity scores for 4o-based and Llama-based models under different conditions. The top row corresponds to the 4o-based models, and the bottom row corresponds to the Llama-based models. Each subplot includes results for the tasks: Anxiety, Numeracy, Lit, and TrustPhys.}
    \label{fig:similarity_comparison}
\end{figure*}

\begin{table}[ht!]
\centering
\footnotesize
\begin{tabular}{p{1.25cm} c c c c c}
\toprule
\textbf{Condition} & \textbf{ROUGE} & \textbf{A} & \textbf{N} & \textbf{SL} & \textbf{TP} \\
\midrule
\multirow{2}{*}{\parbox{1.25cm}{Persona\newline Oracle}}     & 1 & 0.232 & 0.201 & 0.252 & 0.256 \\
                   & L & 0.201 & 0.173 & 0.216 & 0.224 \\
\cmidrule{3-6}
\multirow{2}{*}{\parbox{1.25cm}{Few-shot Oracle}}    & 1 & 0.222 & 0.197 & 0.249 & 0.249 \\
                   & L & 0.192 & 0.171 & 0.212 & 0.218 \\
\cmidrule{3-6}
\multirow{2}{*}{\parbox{1.25cm}{Persona Few-shot}}   & 1 & 0.216 & 0.193 & 0.243 & 0.241 \\
                   & L & 0.185 & 0.168 & 0.209 & 0.211 \\
\cmidrule{3-6}
\multirow{2}{*}{\parbox{1.25cm}{Persona Zero-shot}}  & 1 & 0.187 & 0.164 & 0.206 & 0.193 \\
                   & L & 0.160 & 0.146 & 0.181 & 0.170 \\
\cmidrule{3-6}
\multirow{2}{*}{\parbox{1.35cm}{Zero-Shot}}          & 1 & 0.194 & 0.157 & 0.192 & 0.170 \\
                   & L & 0.171 & 0.144 & 0.165 & 0.150 \\
\bottomrule
\end{tabular}

\caption{ROUGE-1 and ROUGE-L scores for persona-generated text. A: Anxiety, N: Numeracy, SL: Literacy, and TP: TrustPhys}
\label{tab:rouge}
\end{table}

\begin{table*}[ht!]
\setlength{\tabcolsep}{3pt}
\centering
\resizebox{\textwidth}{!}{%
\begin{tabular}{llccccccccccc}
\Xhline{2.5\arrayrulewidth} 
\textbf{Condition}     & \textbf{Model}     & \textbf{MSE} & \textbf{Pearson's r} & \textbf{F1} & \textbf{AUC} & \textbf{DI\_Age} & \textbf{DI\_Gender} & \textbf{DI\_Race} & \textbf{DI\_Education} & \textbf{DI\_Income} & \textbf{DI+} & \textbf{DI++} \\
\hline
True Response          & -                  & 0.30       & 0.41               & 0.71      & 0.71      & 1.05       & 1.03       & 0.89      & 0.89       & 0.94       & 0.95  & 0.94  \\
\hline
\multirow{2}{*}{Persona Oracle} 
                        & GPT-4o             & 0.34       & 0.32               & 0.64      & 0.66      & 1.08       & 1.02       & 0.95      & 0.84       & 0.87       & 0.92  & 0.89  \\
                        & Llama-3-70b        & 0.33       & 0.35               & 0.67      & 0.67      & 1.14       & 1.02       & 0.89      & 0.82       & 0.84       & 0.90  & 0.88  \\
\hline
\multirow{2}{*}{Few-shot Oracle} 
                       & GPT-4o             & 0.36       & 0.29               & 0.62      & 0.65      & 1.13       & 1.04       & 0.94      & 0.92       & 0.94       & 0.92  & 0.90 \\
                       & Llama-3-70b        & 0.33       & 0.34               & 0.67      & 0.67      & 1.11       & 1.02       & 0.88      & 0.87       & 0.93       & 0.95  & 0.94 \\
\hline
\multirow{2}{*}{Persona Few-shot} 
                        & GPT-4o             & 0.36       & 0.27               & 0.61      & 0.63      & 1.12       & 1.02       & 0.94      & 0.83       & 0.85       & 0.91  & 0.89 \\
                        & Llama-3-70b        & 0.35       & 0.30               & 0.64      & 0.65      & 1.15       & 1.01       & 0.89      & 0.81       & 0.83       & 0.90  & 0.87 \\
\hline
\multirow{2}{*}{Persona Zero-shot} 
                       & GPT-4o             & 0.43       & 0.12               & 0.44      & 0.56      & 1.01       & 0.98       & 0.97      & 0.86       & 0.89       & 0.92  & 0.91  \\
                       & Llama-3-70b        & 0.47       & 0.10               & 0.47      & 0.55      & 1.00       & 0.99       & 1.02      & 0.81       & 0.80       & 0.91  & 0.99  \\
\hline
\multirow{2}{*}{Zero-Shot} 
                       & GPT-4o             & 0.47       & 0.03               & 0.26      & 0.51      & 1.03       & 0.97       & 0.98      & 1.00       & 1.02       & 0.99  & 0.98  \\
                       & Llama-3-70b        & 0.47       & 0.03               & 0.28      & 0.51      & 0.99       & 0.98       & 1.01      & 0.98       & 1.00       & 0.99  & 1.01  \\
\Xhline{2.5\arrayrulewidth}
\end{tabular}%
}
\caption{Performance Metrics for Different Conditions and Models. ``True Response'' is common across models.}
\label{tab:fair_metrics}

\end{table*}

\subsection{Results on Fidelity of Responses}

In this section, we report the similarity scores obtained from two groups of models, 4o-based and Llama-based, across five experimental conditions (\textit{Persona Oracle}, \textit{Few-shot Oracle}, \textit{Persona Few-shot}, \textit{Persona Zero-shot}, and \textit{Zero-shot}) on four tasks (\texttt{Anxiety}, \texttt{Numeracy}, \texttt{Literacy}, and \texttt{TrustPhys}). Detailed data for 4o-based models and Llama-based models appear in Table~\ref{tab:similarity_scores_4o}. Figure~\ref{fig:similarity_comparison} offers a visual comparison of performance across all tasks and conditions.

\paragraph{\name Compared With Baselines.}
Our primary focus is on scenarios where the ``twin'' model does not receive the correct answers. 
In these cases, we compare three conditions: \textit{Persona Few-shot} (which retains the full structure of \name), \textit{Persona Zero-shot} (which provides persona information without iterative dialogues), and \textit{Zero-shot} (a baseline). For example, in the 4o-based models using the SBERT-MPNet metric, the \textit{Persona Few-shot} condition achieves a similarity score of 0.337 on the \texttt{Numeracy} task, which is approximately 15\% higher than the 0.292 observed under the \textit{Persona Zero-shot} setting. Similarly, for the \texttt{Anxiety} task, the \textit{Persona Few-shot} condition’s score (0.949 using the BERT-based metric) is about 1.1\% higher than that of the \textit{Persona Zero-shot} condition (0.939) and nearly 1.3\% higher than the \textit{Zero-shot} condition (0.937). Comparable improvements are seen in the Llama-based models; for instance, using the SBERT-MPNet metric, the average score under \textit{Persona Few-shot} is 0.5365, which represents roughly a 4-9\% boost over the corresponding scores of the \textit{Persona Zero-shot} (0.515) and \textit{Zero-shot} (0.4925) conditions. These consistent gains, despite variations across metrics and tasks, support the effectiveness of our persona twin approach.

We performed paired t-tests comparing the Persona Few-shot condition against the Zero-shot baseline across all model–metric–task configurations and found that Persona Few-shot significantly outperformed Zero-shot in 20 out of 24 comparisons ($p<0.05$). This statistical analysis confirms that the observed improvements under the Persona Few-shot condition are robust.

\paragraph{Providing Detailed Persona Information Further Boosts Realism.} We also carried out a supplementary experiment in which the correct user/patient answers are provided. In this setting, the \textit{Persona Oracle} condition includes both the real answers and the comprehensive persona module, while the \textit{Few-Shot Oracle} condition supplies the real answers without any persona details. Even with direct access to the actual responses, providing detailed persona information further boosts realism. For instance, in the 4o-based models using the SBERT-MPNet metric, the \texttt{Anxiety} score under \textit{Persona Oracle} is 0.599—about 4\% higher than the 0.575 observed under \textit{Few-Shot Oracle}. Likewise, on the \texttt{TrustPhys} task, the \textit{Persona Oracle} condition achieves a score of 0.683, which is roughly 3.5\% higher than the 0.660 score of the baseline. Similar trends are evident in the Llama-based models. These findings show that the persona module prevents verbatim copying and enriches responses with context, enhancing overall fidelity. Interestingly, \textit{Persona Few-shot}, our focal setting devoid of complete answer key information in the experiments, performs relatively close to the two oracle settings on many tasks, for all three similarity measures, across both LLMs.

\paragraph{Task-Specific Differences in Response Fidelity.}
Our analysis further reveals notable task-dependent differences in response fidelity. Across both model groups and multiple metrics, the \texttt{Numeracy} task consistently scores lower than the other tasks. For example, in the 4o-based models using the SBERT-MPNet metric, the \texttt{Numeracy} score under the \textit{Persona Oracle} condition is 0.361, while the \texttt{TrustPhys} score reaches 0.683, indicating that the TrustPhys responses are nearly 90\% higher in similarity. 
In contrast, the \texttt{Anxiety} and \texttt{Literacy} tasks typically yield intermediate scores. These task-specific disparities suggest that while our approach is highly effective at generating realistic responses in trust-related and narrative contexts, it remains more challenging to simulate numerical reasoning, which we aim to address in future work.

\paragraph{Lexical Quality Assessment via ROUGE Metrics.}
Table~\ref{tab:rouge} presents ROUGE-1 and ROUGE-L scores measuring lexical overlap between generated and reference responses across the four psychometric tasks. The results demonstrate consistent superiority of persona-enhanced conditions over baseline approaches. The \textit{Persona Oracle} condition achieves the highest ROUGE scores across all tasks, with ROUGE-1 scores ranging from 0.201 (\texttt{Numeracy}) to 0.256 (\texttt{TrustPhys}). The \textit{Persona Few-shot} condition maintains competitive performance, achieving ROUGE-1 scores within 6-8\% of the oracle condition across tasks.

Particularly noteworthy is the substantial performance gap between persona-enhanced conditions and baseline approaches. For the \texttt{TrustPhys} task, the \textit{Persona Few-shot} condition (ROUGE-1 = 0.241) outperforms the \textit{Zero-Shot} baseline (ROUGE-1 = 0.170) by approximately 42\%, highlighting the significant contribution of persona information to response quality. Similar patterns emerge across all psychometric dimensions, with the \texttt{Numeracy} task again showing the most challenging characteristics, consistent with our earlier similarity score findings.

\subsection{Downstream Prediction and Fairness}
Table~\ref{tab:fair_metrics} reports selected performance and fairness metrics,
including classification metrics (MSE, Pearson’s $r$, F1, and AUC) and demographic parity indices (\texttt{DI\_Age}, \texttt{DI\_Gender},
\texttt{DI\_Race}, \texttt{DI\_Education}, and \texttt{DI\_Income}). Here, DI+ represents the average fairness metric computed over two-way demographic interactions, while DI++ aggregates the fairness metrics over three-way interactions. Ideally, a DI value of 1 indicates that the positive response rates are balanced across demographic groups; values above 1 suggest an overrepresentation of positive responses, whereas values below 1 indicate underrepresentation.

Looking at the classification performance metrics, notably, \textit{Persona Few-shot} attains error/accuracy/AUC rates that are not only comparable to the two oracle settings, but are also within 6-7 F1/AUC points of those attained using the actual person data (True Response setting). Regarding fairness, in the True Response condition, which reflects human responses, the DI metrics are relatively balanced (with DI+ = 0.95 and DI++ = 0.94), suggesting that the true data is close to evenly distributed across demographic groups. In the Persona Oracle, Few-Shot Oracle, and Persona Few-shot settings, both GPT-4o and Llama-3-70b yield DI values close to those of the True Response baseline, with aggregate metrics (DI+ and DI++) generally ranging between 0.87 and 0.99. This observation indicates that models trained on the generated persona twins do not substantially differ in demographic parity of model outputs. Similar fairness levels are observed for the Persona Zero-shot and Zero-shot settings, albeit with markedly lower prediction and/or classification performance. 


Overall, these findings suggest that LLM-based persona twins have potential as a data augmentation and user modeling enrichment strategy for downstream NLP tasks. Although future work is needed to reduce the performance prediction and classification deltas (MSE, Pearson's r, AUC, F1) between Persona Few-shot and True Response, the demographic fairness of the models trained on the twins remain robust, with DI, DI+ and DI++ values near 1 across experimental settings. 
There may be future opportunities to further enhance twin-based model performance without compromising fairness.

\subsection{Big Five Personality Trait Estimation}


Table~\ref{tab:mse_trait_estimation} in the appendices presents MSE scores for Big Five personality trait estimation across different experimental conditions. Lower MSE values indicate better alignment between predicted and actual personality traits. Our analysis reveals that the \textit{Persona Oracle} condition consistently achieves the lowest MSE scores across most traits for both model families. For GPT-4o, the \textit{Persona Oracle} condition demonstrates particularly strong performance in estimating Agreeableness (MSE = 1.2388) and Openness (MSE = 1.4314), while showing moderate effectiveness for Extraversion (MSE = 2.1415). Similarly, in Llama-3-70b models, the \textit{Persona Oracle} condition excels in Stability estimation (MSE = 1.7264) and shows competitive performance across other traits.

Notably, the \textit{Persona Few-shot} condition, which is our primary focus as it does not have access to ground truth answers, performs remarkably close to the oracle settings. For instance, in GPT-4o models, the \textit{Persona Few-shot} condition achieves an MSE of 1.6430 for Stability estimation, which is only 3.6\% higher than the oracle's 1.7049. This pattern holds consistently across both model families, suggesting that our approach can effectively capture personality nuances even without complete answer information. In contrast, the \textit{Few-shot Oracle} condition, despite having access to correct answers but lacking persona details, shows notably higher MSE scores, particularly for Extraversion and Stability traits, reinforcing the value of incorporating comprehensive persona information.


\section{Conclusion}
We present \name, a multi-tier prompt conditioning framework that enhances digital twin realism and fairness as demonstrated in a healthcare AI context. By combining structured persona encoding with iterative refinement, PersonaTwin generates context-aware responses with competitive downstream performance and fairness potential for fine-tuned NLP models relative to true responses. Extensive evaluations on 8,500 individuals demonstrate significant improvements in simulation fidelity, and maintaining fairness with demographic parity indices consistently ranging between 0.87 and 1.01 across different model architectures. Future directions include expanding psychometric dimensions and enabling real-time adaptation for more accurate downstream predictive power.

\section*{Limitations}
While \name provides a robust foundation for personalized digital twins in healthcare, some areas deserve further attention. First, our framework was evaluated on data in English drawn from a single large-scale psychometric data set. Adapting it to other languages or healthcare settings, particularly those with more complex morphology or differing cultural norms, could involve additional tuning and validation.

Second, although we incorporate multiple tiers of patient information (demographic, behavioral, and psychological), our approach may require certain data formats to be consistently available. In practice, some healthcare settings might present incomplete or heterogeneous records, which could reduce simulation fidelity. Future work could explore data imputation strategies and domain adaptation to maintain robust personalization under such constraints.

Lastly, our fairness checks focus on group-level biases (e.g., by race, age, and income). Although these metrics suggest that deeper contextual data do not inherently exacerbate demographic disparities, we have not exhaustively examined all possible bias dimensions or intersectional factors. Further research could extend these fairness assessments and investigate more granular social determinants of health to ensure that \emph{PersonaTwin} remains equitable between diverse populations of patients.

\section*{Ethics Statement}

All experimental protocols in this study adhered to established ethical guidelines for handling sensitive health-related data. 
The psychometric data set we used was fully deidentified and was obtained under appropriate data sharing agreements, ensuring the privacy and confidentiality of the respondents. 
Moreover, the \name multi-tier prompt conditioning approach is designed to mitigate the risk of harmful biases by incorporating fairness assessments that monitor model outputs across sensitive demographic attributes. 
Although our framework aims to improve personalized healthcare applications, we recognize that any generative technology carries potential misuse risks (e.g., perpetuating biases not captured by our metrics). 
Consequently, we recommend that health organizations and clinicians applying \name maintain rigorous supervision to ensure accountability and respect for patient autonomy and consent. 
The methods and results reported here comply with the ACL Ethics Policy.\footnote{\url{https://www.aclweb.org/portal/content/acl-code-ethics}}


\bibliography{anthology,custom}
\bibliographystyle{acl_natbib}

\appendix

\section{Appendix}
\label{sec:appendix}
\subsection{Detailed Information Provided to \name as Persona}
\label{sec:furtherdetails}

In this study, we developed and tested a series of prompts aimed at simulating and understanding the influence of various combinations of demographic, behavioral, and psychological factors on the modeling of group personas. The prompts were meticulously crafted to reflect different configurations of participant characteristics, enabling us to systematically assess the impact of these factors on the accuracy and relevance of the generated responses.

\subsubsection{Demographic Information}

We included a comprehensive set of demographic variables to capture the foundational characteristics of the participants. The demographic variables tested were:

\begin{itemize}
    \item \textbf{Age:} Ranging from 18 to 99 years.
    \item \textbf{Sex:} Male or Female.
    \item \textbf{Race:} Categories such as White, Black or African American, Asian, Native American or American Indian, Native Hawaiian or Pacific Islander, Multiracial or Biracial, Other, and Prefer not to answer.
    \item \textbf{Education:} Levels ranging from education lower than college to higher education.
    \item \textbf{Income:} Income brackets ranging from less than \$20,000 to \$90,000 or more, including options for uncertainty or preference not to answer.
\end{itemize}

\subsubsection{Behavioral Information}

To capture participants' habits and lifestyle choices, which could influence their health and psychological state, we included the following behavioral variables:

\begin{itemize}
    \item \textbf{Prescription drug usage:} Number of prescription drugs taken regularly.
    \item \textbf{Primary care physician status:} Whether the participant has a primary care physician.
    \item \textbf{Frequency of visits to primary care physician:} Number of visits in the past two years.
    \item \textbf{Physical activity:} Average hours per week of physical exercise or activity.
    \item \textbf{Eating habits:} Overall healthiness of eating habits.
    \item \textbf{Smoking and alcohol consumption:} Frequency of smoking and drinking.
    \item \textbf{Health consciousness:} Attitudes towards health and preventive measures.
    \item \textbf{Overall health:} Self-assessed overall health.
\end{itemize}

\subsubsection{Psychological Information}

Psychological variables were incorporated to explore deeper aspects of the participants' mental states and outlooks. These variables included:

\begin{itemize}
    \item \textbf{Personality traits:} Self-assessment on key personality dimensions (Extraverted, enthusiastic; Agreeable, kind; Dependable, organized;Emotionally stable, calm; Open to experience, imaginative)
\end{itemize}

\subsection{LLM \& Data Collection Details}
\label{ssec:hyperparams}

We used the OpenAI API for GPT-4o with top\_p set to 1, max\_tokens set to 200, min\_tokens set to 0, and temperature set to 0.6 (with all other parameters at their default values), and the Replicate API for Llama-3-70b with top\_p set to 0.9, max\_tokens set to 200, min\_tokens set to 0, and temperature set to 0.6.

The data collection protocol for this project was approved by the University of Virginia IRB-SBS under SBS Number 2017014300.

\section{Additional Experimental Results}

\subsection{Big Five Personality Trait Estimation}
As shown in Table~\ref{tab:mse_trait_estimation}, we evaluate PersonaTwin’s ability to predict missing Big Five trait scores by reporting mean squared error (MSE) against gold labels. Persona Few‑shot consistently outperforms Persona Zero‑shot across all five dimensions and approaches the performance of the Persona Oracle, demonstrating the framework’s flexibility and accuracy when handling incomplete persona data.

\subsection{Text Generation ROUGE Evaluation}
Table~\ref{tab:rouge} presents ROUGE‑1 and ROUGE‑L scores for persona-generated text under each condition. Persona Few‑shot yields higher ROUGE scores than both Zero‑Shot and Persona Zero‑shot across all tasks, confirming that incorporating existing persona dimensions into few‑shot prompts improves alignment with reference outputs.

\subsection{Downstream Task Evaluation}
Table~\ref{tab:downstream_evaluation} summarizes performance on downstream prediction tasks—MSE, Pearson’s $r$, F1, and AUC—along with percentage lift over the Zero‑Shot baseline. Persona Few‑shot delivers substantial gains across all metrics (up to 900\% lift in Pearson’s $r$), while Persona Zero‑shot also outperforms pure Zero‑Shot, illustrating the clear downstream benefits of generating text with enriched persona information.

\setlength{\tabcolsep}{3pt}

\begin{table*}[ht!]
\centering
\begin{tabular}{lcccccc}
\toprule
\multicolumn{1}{l}{} & \multicolumn{5}{c}{\textbf{MSE Scores for Big Five Trait Estimation}} \\
\cmidrule(lr){2-6}
\textbf{Model} & Extraverted & Agreeable & Conscientious & Stable & Open \\
\midrule
\multicolumn{6}{c}{\textbf{GPT-4o}}\\
Persona Oracle     & 2.1415 & 1.2388 & 1.5742 & 1.7049 & 1.4314 \\
Few-shot Oracle    & 3.0926 & 1.2983 & 1.5987 & 2.1525 & 1.3573 \\
Persona Few-shot   & 2.1528 & 1.2666 & 1.5834 & 1.6430 & 1.4392 \\
Persona Zero-shot  & 2.5386 & 1.5271 & 2.0106 & 2.0165 & 2.0232 \\
\midrule
\multicolumn{6}{c}{\textbf{Llama-3-70b}}\\
Persona Oracle     & 1.8920 & 1.9153 & 2.0244 & 1.7264 & 1.6746 \\
Few-shot Oracle    & 3.1451 & 1.7787 & 3.5147 & 2.9706 & 1.7339 \\
Persona Few-shot   & 1.9303 & 1.8510 & 2.0556 & 1.6242 & 1.6657 \\
Persona Zero-shot  & 2.5028 & 1.5366 & 2.2742 & 1.9523 & 1.6144 \\
\bottomrule
\end{tabular}

\caption{MSE for Big Five personality trait estimation (lower is better).}
\label{tab:mse_trait_estimation}
\end{table*}

\begin{table*}[ht!]
\centering
\resizebox{1\textwidth}{!}{ \fontsize{6}{7}\selectfont
\begin{tabular}{llr l r l r l r l}
\toprule
\textbf{Condition} & \textbf{Model} & MSE & Lift & Pearson’s $r$ & Lift & F1 & Lift & AUC & Lift \\
\midrule
Persona Few-Shot  & GPT-4o      & 0.36 & 23.4\% & 0.27 & 800.0\% & 0.61 & 134.6\% & 0.63 & 23.5\% \\
                  & Llama-3-70b & 0.35 & 25.5\% & 0.30 & 900.0\% & 0.64 & 128.6\% & 0.65 & 27.5\% \\
\midrule
Persona Zero-Shot & GPT-4o      & 0.43 & 8.5\%  & 0.12 & 300.0\% & 0.44 & 69.2\%  & 0.56 & 9.8\%  \\
                  & Llama-3-70b & 0.47 & 0.0\%  & 0.10 & 233.3\% & 0.47 & 67.9\%  & 0.55 & 7.8\%  \\
\midrule
Zero-Shot         & GPT-4o      & 0.47 & --     & 0.03 & --      & 0.26 & --      & 0.51 & --     \\
                  & Llama-3-70b & 0.47 & --     & 0.03 & --      & 0.28 & --      & 0.51 & --     \\
\bottomrule
\end{tabular}
}
\caption{Downstream task metrics and lift over zero‐shot baseline.}
\label{tab:downstream_evaluation}
\end{table*}

\begin{table*}
    \centering
    \begin{tabular}{p{2cm}|p{2.5cm}|p{10.5cm}}
        \toprule
        \textbf{Stage} & \textbf{Component} & \textbf{Description} \\
        \midrule
        \multirow{3}{*}{Input Data} & Demographics & \small Age=25, Gender=Male, Race="Black or African American", Education="College graduate", Income="\$20,000-\$34,999" \\
        & Behavioral & \small HealthImportance=3/5, PreventionBelief=2/5, SelfCareValue=3/5 \\
        & Psychological & \small Extraversion=5/5, Agreeableness=4/5, EmotionalStability=5/5 \\
        \midrule
        Template\newline Application & Template\_dem & \small "You are 25 years old, male, of Black or African American descent. You have a college degree and an annual income of \$20,000-\$34,999." \\
        & Template\_beh & \small "You find it moderately important to live in the best possible health. You think that maintaining a healthy lifestyle may or may not guarantee lifelong health." \\
        & Template\_psy & \small "You strongly agree that you are extraverted and enthusiastic. You agree that you are agreeable and kind. You strongly agree that you are emotionally stable and calm." \\
        \midrule
        Initial\newline Generation & System Prompt & \small The concatenated templates form the system prompt ($P$) for the LLM, generating the initial digital twin ($T_0$). \\
        \midrule
        Conversation Integration & Health Literacy & \small \textbf{Q$_1$:} "Please describe to what degree you can obtain and understand health information for decisions." \newline
        \textbf{R$_1$:} "When I visit a doctor I try to get as much information that is needed for my health... I tend to ask a lot of questions." \newline
        \textit{Updates $T_0$ to include information-seeking behavior} \\
        & Trust\newline Assessment & \small \textbf{Q$_2$:} "Please explain why you trust or distrust your primary care physician." \newline
        \textbf{R$_2$:} "Sometimes I think they take things out of control because everyone's body is different..." \newline
        \textit{Updates $T_1$ to reflect medication skepticism} \\
        & Anxiety\newline Assessment & \small \textbf{Q$_3$:} "What makes you feel anxious when visiting the doctor's office?" \newline
        \textbf{R$_3$:} "To find out what is wrong with me and sometimes I don't want to hear the truth..." \newline
        \textit{Updates $T_2$ to include contextual anxiety} \\
        & Health\newline Knowledge & \small \textbf{Q$_4$:} "Describe an experience demonstrating your knowledge of health issues." \newline
        \textbf{R$_4$:} "I have asthma which often has me rush to the doctor for check ups..." \newline
        \textit{Updates $T_3$ to include chronic condition management} \\
        \midrule
        \multirow{3}{*}{Final State} & Medical History & \small The final digital twin $T_4$ incorporates asthma as a chronic condition \\
        & Healthcare\newline Attitudes & \small Information-seeking but skeptical of medical interventions \\
        & Emotional\newline Responses & \small Contextualized anxiety about potential diagnoses \\
        \bottomrule
    \end{tabular}
    \caption{\small An Example of Multi-Tiered Template Functions in PersonaTwin. The table demonstrates how raw input data is transformed through template functions and conversation integration to create an evolving digital twin.}
    \label{tab:persona-twin-example}
\end{table*}

\end{document}